\documentclass[letterpaper]{article} 

\usepackage{aaai23}

\usepackage{times}  \usepackage{helvet}  \usepackage{courier}  \usepackage[hyphens]{url}  \usepackage{graphicx}    \usepackage{natbib}  \usepackage{caption}     \usepackage{algorithm}
\usepackage{algorithmic}

\usepackage{newfloat}
\usepackage{listings}

\usepackage{booktabs}
\usepackage{array}
\usepackage{multirow}
\usepackage{multicol}
\usepackage{arydshln}
\usepackage{amssymb}
\usepackage{amsmath}
\usepackage{xspace}
\usepackage{bm}
\usepackage{bbm}
\usepackage{adjustbox}
\usepackage{soul}
\usepackage{tikz}
\usetikzlibrary{shapes.geometric, arrows, positioning, calc}
\usepackage{afterpage}
\usepackage{rotating}
\usepackage[normalem]{ulem}

\usepackage{hyperref}
\hypersetup{
    colorlinks=true,
    citecolor=.,
    urlcolor=blue,
    linkcolor=.,
}

\newcommand\drc{\textsc{DRC}\xspace}
\newcommand\dataset{\textsc{QASPER}\xspace}
\newcommand\dpr{\textsc{DPR}\xspace}
\newcommand\dprft{\textsc{DPR-ft}\xspace}
\newcommand\electra{\textsc{ELECTRA}\xspace}
\newcommand\electrace{\textsc{ELECTRA\_CE}\xspace}
\newcommand\electraceft{\textsc{ELECTRA\_CE-ft}\xspace}
\newcommand\led{\textsc{LED}{}\xspace}
\newcommand\bestmatch{\textsc{BM25}\xspace}
\newcommand\dit{\textsc{DiT}\xspace}
\newcommand\unifiedqa{\textsc{UnifiedQA}\xspace}
\newcommand\unifiedqaft{\textsc{UnifiedQA-ft}\xspace}
\newcommand\pdfminer{\emph{pdfminer.six}\xspace}
\newcommand\pdftoimage{\emph{pdf2image}\xspace}
\newcommand\tesseract{\emph{TesseractOCR}\xspace}
\newcommand\answerf{Answer-$F_1$\xspace}
\newcommand\evidencef{Evidence-$F_1$\xspace}
\newcommand\ditt{\textsc{DiT}}
\newcommand\pdfminert{\emph{pdfminer.six}*}

\newcommand{\set}[1]{\mathcal{#1}}
\newcommand{\sectionsymb}{\S}

\DeclareCaptionStyle{ruled}{labelfont=normalfont,labelsep=colon,strut=off} 
\floatstyle{ruled}
\newfloat{listing}{tb}{lst}{}
\floatname{listing}{Listing}
\title{Detect, Retrieve, Comprehend: A Flexible Framework for Zero-Shot Document-Level Question Answering}
\author{
Tavish McDonald\textsuperscript{\rm 1},
    Brian Tsan \textsuperscript{\rm 2},
    Amar Saini \textsuperscript{\rm 1},
    Juanita Ordonez \textsuperscript{\rm 1}, \\
    Luis Gutierrez \textsuperscript{\rm 1},
    Phan Nguyen \textsuperscript{\rm 1},
    Blake Mason \textsuperscript{\rm 1},
    Brenda Ng \textsuperscript{\rm 1}
}
\affiliations{
\textsuperscript{\rm 1}Lawrence Livermore National Laboratory;
    \textsuperscript{\rm 2}University of California, Merced;\\

\{mcdonald53, saini5, ordonez2, gutierrez74, nguyen97, mason35, ng30\}@llnl.gov;
    btsan@ucmerced.edu\\
}

\newcommand\Tstrut{\rule{0pt}{2.6ex}}
\newcommand\Bstrut{\rule[-0.9ex]{0pt}{0pt}}   
\begin{document}
\maketitle

\begin{abstract}
Researchers produce thousands of scholarly documents containing valuable technical knowledge. The community faces the laborious task of reading these documents to identify, extract, and synthesize information. To automate information gathering, document-level question answering (QA) offers a flexible framework where human-posed questions can be adapted to extract diverse knowledge. Finetuning QA systems requires access to labeled data (tuples of context, question and answer). However, data curation for document QA is uniquely challenging because the context (i.e., text passage containing evidence to answer the question) needs to be \emph{retrieved} from potentially long, ill-formatted documents. Existing QA datasets sidestep this challenge by providing short, well-defined contexts that are unrealistic in real-world applications. We present a three-stage document QA approach: (1) text extraction from PDF; (2) evidence retrieval from extracted texts to form well-posed contexts; (3) QA to extract knowledge from contexts to return high-quality answers -- extractive, abstractive, or Boolean. Using the \dataset dataset for evaluation, our \emph{Detect-Retrieve-Comprehend} (\drc) system achieves a +7.19 improvement in \answerf over existing baselines due to superior context selection. Our results demonstrate that \drc holds tremendous promise as a flexible framework for practical scientific document QA. 
\end{abstract} 

\section{Introduction}
Growth in new machine learning publications has exploded in recent years, with much of this activity occurring outside traditional publication venues. For example, arXiv hosts researchers’ manuscripts detailing the latest progress and burgeoning initiatives. In 2021 alone, over 68,000 machine learning papers were submitted to arXiv. Since 2015, submissions to this category have increased yearly at an average rate of 52\%. While it is admirable that the accelerated pace of AI research has produced many innovative works and manuscripts, the sheer amount of papers makes it prohibitively difficult to keep pace with the latest developments in the field. Increasingly, researchers turn to scientific search engines (e.g., Semantic Scholar and Zeta Alpha), powered by neural information retrieval, to find relevant literature. To date, scientific search engines  \cite{fadaee2020Search,Zhaolee2020SOCO,Parisotzavrel2022Multi} have focused on serving recommendations based on semantic similarity and lexical matching between a query phrase and a collection of document-derived contents, particularly titles and abstracts. Other efforts to elicit the details of scholarly papers have extracted quantified experimental results from structured tables \cite{Kardas2020axcell} and generated detailed summaries from the hierarchical content of scientific documents \cite{Sotudeh2020Generating}. 

\begin{figure}[!t] 
\centering
\usetikzlibrary{decorations.pathreplacing}

\tikzstyle{doc_node}  = [rectangle, fill=white!10]
\tikzstyle{caption_node}  = [rectangle, fill=white, minimum width=1cm, minimum height=3mm, text width=20mm,font={\fontsize{6pt}{6pt}\selectfont}]

\tikzstyle{question_node}  = [rectangle, dotted, draw=black, fill=white, minimum width=1cm, minimum height=5mm, text width=40mm,font=\qcrfamily\fontsize{5pt}{5pt}\selectfont,align=justify]
\tikzstyle{evidence_node}  = [rectangle, dotted, draw=black, fill=white, minimum width=1cm, minimum height=7mm, text width=40mm,font=\qcrfamily\fontsize{5pt}{5pt}\selectfont,align=justify]
\tikzstyle{answer_node}  = [rectangle, dotted, draw=black, fill=white, minimum width=1cm, minimum height=7mm, text width=40mm,font=\qcrfamily\fontsize{5pt}{5pt}\selectfont,align=justify]

\tikzstyle{zoomline}  = [line width=0.2mm,dotted, dash pattern=on 0 off 1cm/20,line cap=rect]
\tikzstyle{line}  = [line width=0.3mm]
\tikzstyle{arrow} = [thick,<-,>=stealth,line width=0.3mm]

\newcommand{\qcrfamily}{\fontfamily{qcr}\fontdimen2\font=0.4em  \fontdimen3\font=0.2em  \fontdimen4\font=0.1em  \fontdimen7\font=0.1em  \hyphenchar\font=`\-    }
\newcommand*{\highlight}[2]{\tikz[baseline=(X.base)] \node[rectangle, fill=#1, inner sep=0.3mm, text width=39mm] (X) {#2};}

\begin{adjustbox}{scale={0.75}{0.75}, center}
    \begin{tikzpicture}[node distance=2cm]
    \fontfamily{cmss}{\fontsize{7}{7}\selectfont
    
    \node (pdf_file) [doc_node] { \includegraphics[page=1,width=4cm]{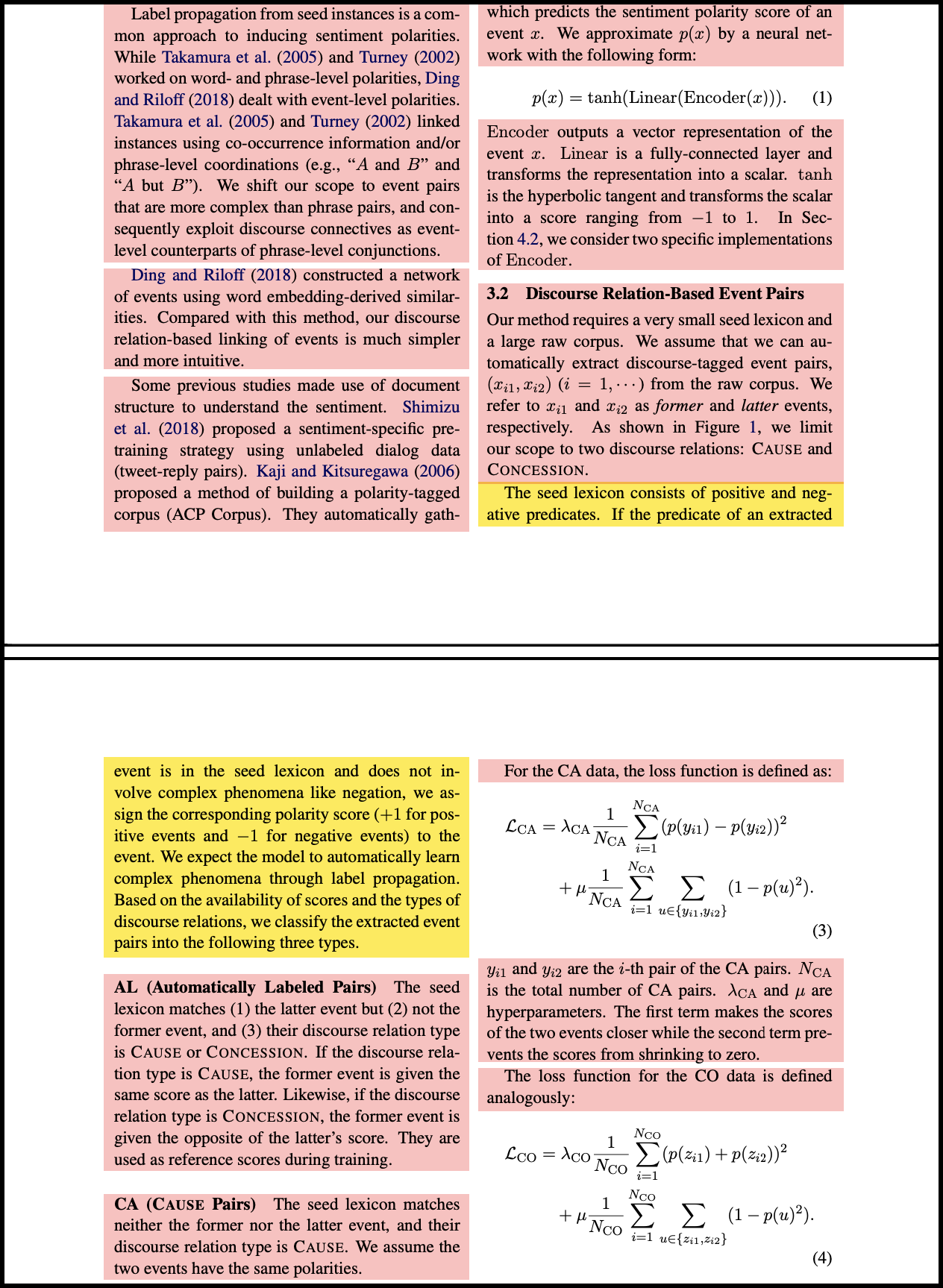} };
    \node (question) [question_node, anchor=north west, below right=5mm and 8mm of pdf_file.north east,label={[shift={(-1.6,0)}]above:Question}] {
        \highlight{cyan!20} {
            What is the seed lexicon?
        }
    };
    \node (evidence) [evidence_node, anchor=north, below=4mm of question.south, label={[shift={(-1.6,0)}]above:Evidence}] {
        \highlight{yellow!80} {
            The seed lexicon consists of positive and negative predicates. If the predicate of an extracted event is in the seed lexicon and does not involve complex phenomena like negation, we assign the corresponding polarity score (+1 for positive events and -1 for negative events) to the event.
        We expect the model to automatically learn complex phenomena through label propagation. Based on the availability of scores and the types of discourse relations, we classify the extracted event pairs into the following three types.
        }
    };
\draw [zoomline] (evidence.north west) ++(0,0) -- ++(-2.86,-0.55);
    \draw [zoomline] (evidence.north west) ++(0,-0.35) -- ++(-1.34,-0.195);
    \draw [zoomline] (evidence.south west) ++(0,0) -- ++(-4.45,0.27);
    \draw [zoomline] (evidence.south west) ++(0,0.11) -- ++(-2.9,0.16);
    \node (answer) [answer_node, anchor=north, below=4mm of evidence.south, label={[shift={(-1.66,0)}]above:Answer}] {
        \highlight{green!20} {
            a vocabulary of positive and negative predicates that helps determine the polarity score of an event
        }
    };
\draw [decorate, decoration={brace},line] (evidence.east) ++(0.1,1.38) -- ++(0,-1.48);
    \draw [arrow] (answer.east) ++(0,0) -- ++(0.6,0) -- ++(0,2.9) -- ++(-0.4,0);
    \draw [line] (evidence.east) ++(0.1,1.38) -- ++(-4.4,0);
    \draw [line] (evidence.east) ++(0.1,-0.09) -- ++(-4.4,0);
    \draw [line] (evidence.east) ++(0.1,-0.09) ++(-4.4,0) -- ++(0,1.48);
    }
\end{tikzpicture}
\end{adjustbox}

\colorlet{soulblue}{cyan!20}
\colorlet{soulyellow}{yellow!80}
\colorlet{soulgreen}{red!20}
\colorlet{soulred}{green!20}

\DeclareRobustCommand{\hlblue}[1]{\sethlcolor{soulblue}\hl{#1}}
\DeclareRobustCommand{\hlyellow}[1]{\sethlcolor{soulyellow}\hl{#1}}
\DeclareRobustCommand{\hlred}[1]{\sethlcolor{soulgreen}\hl{#1}}
\DeclareRobustCommand{\hlgreen}[1]{\sethlcolor{soulred}\hl{#1}}

\caption{%
    \dataset \hlblue{questions} require \hlred{PDF text extraction} and \hlyellow{evidence} retrieval to generate an \hlgreen{answer}.
}

\label{fig:dataset_diagram}
\end{figure}
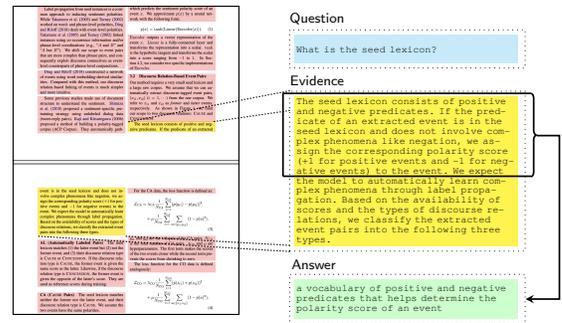 

\begin{figure*}[htp]

\newcommand{\qcrfamily}{\fontfamily{qcr}\fontdimen2\font=0.4em  \fontdimen3\font=0.2em  \fontdimen4\font=0.1em  \fontdimen7\font=0.1em  \hyphenchar\font=`\-    }

\tikzstyle{doc_node}  = [rectangle, draw=black, fill=white!10, minimum width=1cm, minimum height=7mm, text centered, text width=16mm]
\tikzstyle{pdf_label}  = [rectangle, draw=red, fill=red, text=white, minimum width=1cm, minimum height=5mm, text centered, text width=16mm]
\tikzstyle{text_node}  = [rectangle, draw=black, fill=white!10, minimum width=1cm, minimum height=7mm, text width=22mm,font=\qcrfamily\fontsize{4pt}{4pt}\selectfont,align=justify]
\tikzstyle{question_node}  = [rectangle, draw=black, fill=white!40, minimum width=2cm, minimum height=7mm, text width=20mm,font=\qcrfamily\fontsize{5pt}{5pt}\selectfont,align=justify]
\tikzstyle{k_text_node}  = [rectangle, draw=black, fill=white!10, minimum width=1cm, minimum height=13mm, text width=22mm,font=\qcrfamily\fontsize{4pt}{4pt}\selectfont,align=justify]
\tikzstyle{rank_node}  = [rectangle, draw=black, fill=red!20, minimum width=8mm, minimum height=12mm, text width=8mm,font=\qcrfamily\fontsize{5pt}{5pt}\selectfont,align=justify,text depth=10.2mm]
\tikzstyle{code_node}  = [rectangle, draw=black, rounded corners=0.5mm, fill=red!20, minimum width=1cm, minimum height=7mm, text centered, text width=16mm]
\tikzstyle{model_node} = [rectangle, draw=black, rounded corners=0.5mm, fill=orange!20, minimum width=1cm, minimum height=1cm, text centered, text width=16mm]
\tikzstyle{arrow} = [thick,->,>=stealth]
\tikzstyle{box_label} = [line width=0.01mm,font=\fontfamily{lmss}\fontsize{10pt}{10pt}\selectfont]

\begin{adjustbox}{scale={0.75}{0.75}, center}
\begin{tikzpicture}[node distance=2cm]
    
\fontfamily{cmss}{\fontsize{7}{7}\selectfont
    \node (detect_box) [above, fill=cyan!10, minimum width=18cm, minimum height=4cm] at (7.8,-1.8){};
    \node (retrieve_box)   [above, dotted,fill=green!10, minimum width=8.8cm, minimum height=3.6cm] at (3.2,-5.7){};
    \node (comprehend_box) [above, fill=blue!10,  minimum width=8.8cm, minimum height=3.6cm] at (12.4,-5.7){};
    \node[box_label, anchor=north west, below right=0mm and 0mm of detect_box.north west] {\textbf{\textit{Detect} (Text Extraction)}};
    \node[box_label, anchor=south west, above right=0mm and 0mm of retrieve_box.south west] {\textbf{\textit{Retrieve} (Evidence Retrieval)}};
    \node[box_label, anchor=south east, above left=0mm and 0mm of comprehend_box.south east] {\textbf{\textit{Comprehend} (Question Answering)}};
\node (pdf_file) [doc_node,label={[shift={(0,-0.15)}]below:PDF File}] { \includegraphics[page=2,width=1.61cm]{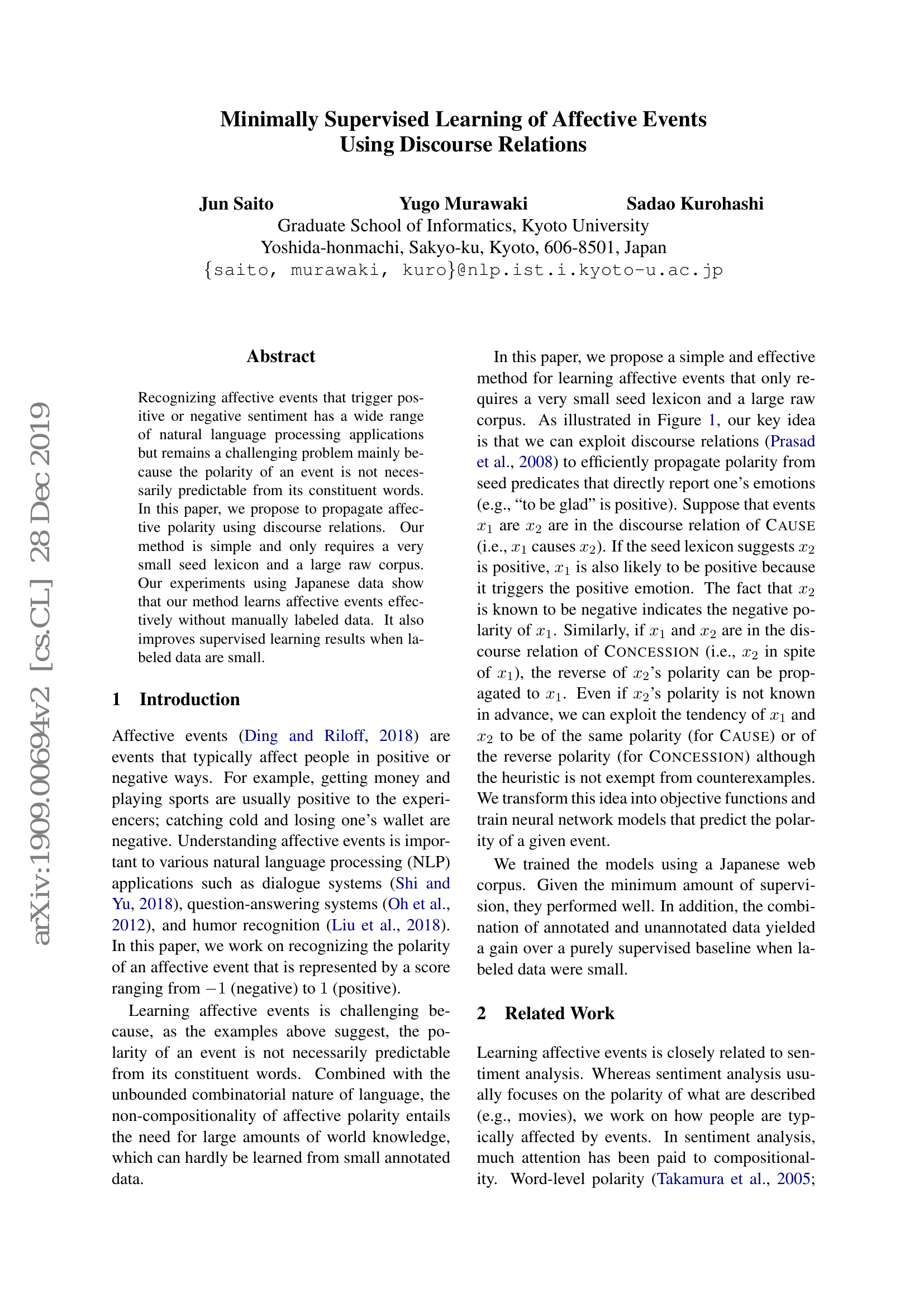} };
    \node (pdf_file_banner) [pdf_label] {PDF};
    \node (pdf2image) [code_node, right=6mm of pdf_file] {\pdftoimage};
    \node (pdf_images_1) [doc_node, right=6mm of pdf2image,label={[shift={(0.15,-0.2)}]below:Page Images}] { \includegraphics[page=2,width=1.61cm]{Figures/1909.00694.page_1.png} };
    \node (dit) [model_node, right=7mm of pdf_images_1] {\dit};
    \node (bounding_box_1) [doc_node, right=6mm of dit,label={[shift={(0.15,-0.2)}]below:Paragraph Bounding Boxes}] { \includegraphics[page=2,width=1.61cm]{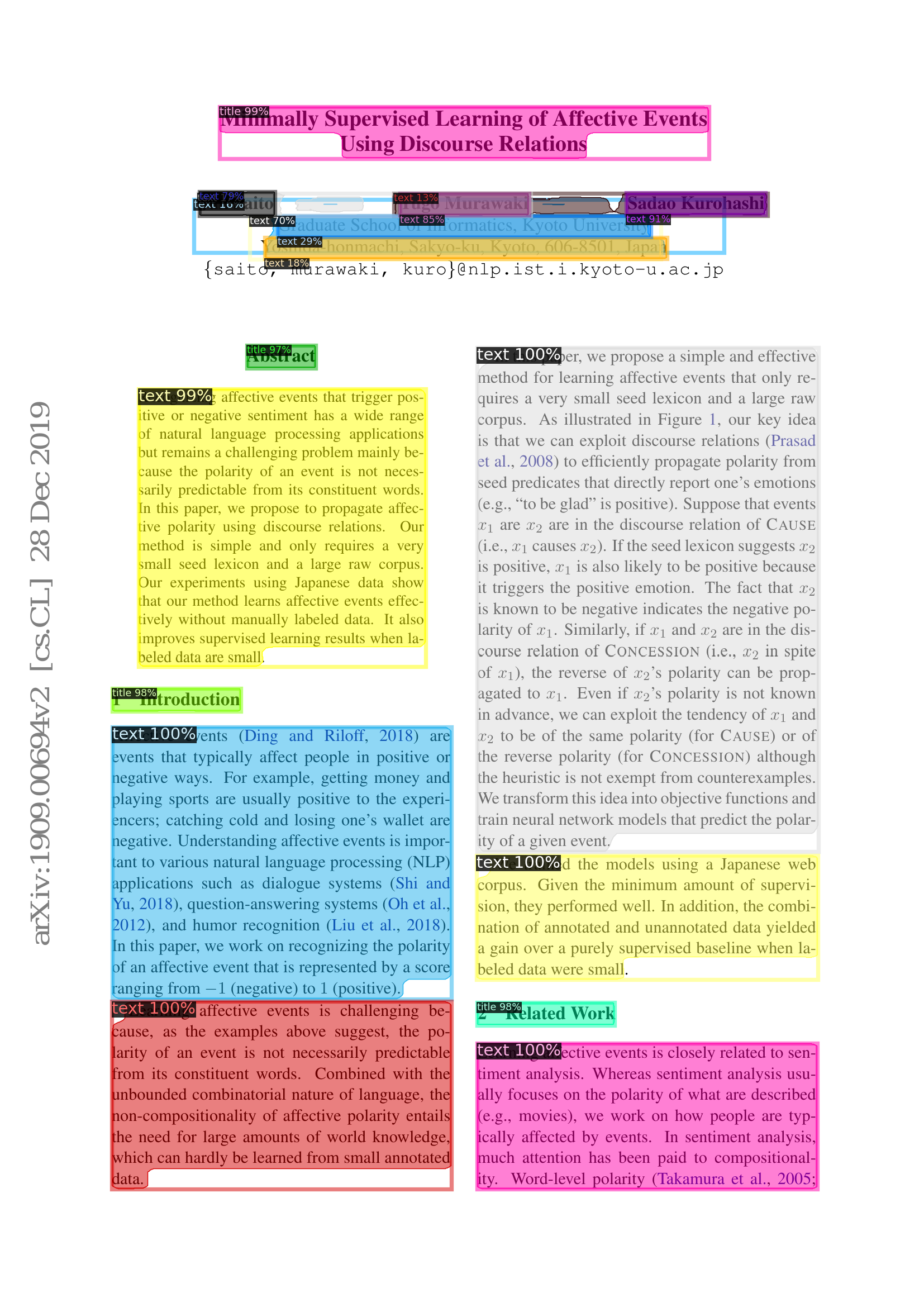} };
    \node (pdfminer) [code_node, right=8mm of bounding_box_1] {\pdfminer};
    \node (paragraph_text_1) [text_node, above right=-10mm and 6mm of pdfminer,label={[shift={(0.25,-0.35)}]below:Paragraph Texts}] {
        CO (CONCESSION Pairs) The seed lexicon matches neither the former nor the latter event, and their discourse relation type is CONCESSION. We assume the two events have the reversed polarities.
    };
    \foreach \X [count=\Y] in {2,...,5} {
        \node (paragraph_text_\X) [text_node, anchor=north west,below right=0.8mm and 0.8mm of paragraph_text_\Y.north west] {
            CO (CONCESSION Pairs) The seed lexicon matches neither the former nor the latter event, and their discourse relation type is CONCESSION. We assume the two events have the reversed polarities.};
    }
\node (question) [question_node, anchor=north, below right=20mm and 5mm of pdf_file.south,label={[shift={(0.0,0.0)}]above:Question Text}] {What is the seed lexicon?};

    \node (crossencoder) [model_node, right=6mm of question, text width=18mm] {\electrace};
    
\node (k_paragraph_text_1) [k_text_node, above right=-10mm and 6mm of crossencoder,label={[shift={(0.5,1.75)}]below:Top-K Paragraph Texts (K=3)}] {
        CO (CONCESSION Pairs) The seed lexicon matches neither the former nor the latter event, and their discourse relation type is CONCESSION. We assume the two events have the reversed polarities.
    };
    \node (k_paragraph_rank_1) [rank_node, anchor=west, right=-0.5mm of k_paragraph_text_1] {r=0.07};
    
\node (k_paragraph_text_2) [k_text_node, anchor=north west,below right=3mm and 2mm of k_paragraph_text_1.north west] {
        CA (CAUSE Pairs) The seed lexicon matches neither the former nor the latter event, and their discourse relation type is CAUSE. We assume the two events have the same polarities.
    };
    \node (k_paragraph_rank_2) [rank_node, anchor=west, right=-0.5mm of k_paragraph_text_2] {r=0.10};
    \node (unifiedqa) [model_node, right=18mm of k_paragraph_text_2] {\unifiedqa};
    \node (k_answers_1) [question_node, anchor=center,below right=-7mm and 25mm of unifiedqa.north west, label={[shift={(0.0,0)}]above:K Answers (K=3)}] {no answer};
\node (k_answers_2) [question_node, anchor=center,right=25mm of unifiedqa.west] {matches neither the former nor the latter event};
\node (k_answers_3) [question_node, anchor=center,below right=10mm and 25mm of unifiedqa.north west] {positive and negative predicates};

\draw [arrow] (pdf_file) -- (pdf2image);
    \draw [arrow] (pdf2image) -- (pdf_images_1);
    \draw [arrow] (pdf_images_1) -- (dit);
    \draw [arrow] (dit) -- (bounding_box_1);
    \draw [arrow] (bounding_box_1) -- (pdfminer);
    \draw [arrow] (pdfminer) -- ++(1.48,0) (paragraph_text_1);
    \draw [arrow] (pdf_file) - ++(0,1.5) -| (pdfminer);
\draw [arrow] (paragraph_text_5.south) ++(-0.2,-0.38) -- ++(0,-0.6) -| (crossencoder);
    \draw [arrow] (question) -- (crossencoder);
    \draw [arrow] (crossencoder) -- ++(1.58,0) (k_paragraph_text_2);
    \draw [arrow] (k_paragraph_rank_2) -- (unifiedqa);
\draw [arrow] (question) -- ++(0,-1.35) -| (unifiedqa);
    \draw [arrow] (unifiedqa) -- ++(1.2,0) |- (k_answers_1);
    \draw [arrow] (unifiedqa) -- (k_answers_2);
    \draw [arrow] (unifiedqa) -- ++(1.2,0) |- (k_answers_3);

\node (k_paragraph_text_3) [k_text_node, anchor=north west,below right=3mm and 2mm of k_paragraph_text_2.north west] { 
        The seed lexicon consists of positive and negative predicates. If the predicate of an extracted
    };
            
\node (k_paragraph_rank_3) [rank_node, anchor=west, right=-0.5mm of k_paragraph_text_3] {r=0.93};
    
\foreach \X [count=\Y] in {2,...,3} {
        \node (pdf_images_\X) [doc_node, anchor=north west,below right=0.8mm and 0.8mm of pdf_images_\Y.north west] { \includegraphics[page=2,width=1.61cm]{Figures/1909.00694.page_1.png} };
    }
\foreach \X [count=\Y] in {2,...,3} {
        \node (bounding_box_\X) [doc_node, anchor=north west,below right=0.8mm and 0.8mm of bounding_box_\Y.north west] { \includegraphics[page=2,width=1.61cm]{Figures/1909.00694.page_1_bboxes.png} };
    }
    
    }
    \end{tikzpicture}
\end{adjustbox}
\caption{An instance of our modular end-to-end \drc system comprised of \dit{ + }\electrace{ + }\unifiedqa.}
\label{fig:system_arch}
\end{figure*}

While these scientific search engines suffice for topic exploration, once a set of papers are identified as relevant, researchers would want to probe deeper for information to address specific questions conditioned on their prior domain knowledge (e.g., \emph{What baselines is the neural relation extractor compared to?}). While one can gain a sense of the main findings of a paper by reading the abstract, the answers to these probing questions are frequently found in the details of the methodology, experimental setup, and results sections. Furthermore, questions may require synthesis of document passages to produce an abstractive answer rather than simply extracting a contiguous span. Reading and manually cross-referencing the results of several papers is a labor-intensive approach to glean specific knowledge from scientific documents. Therefore, effective tools to help automate knowledge discovery are sorely needed. 

A promising approach to extracting knowledge from scientific publications is document-level question answering (QA): using an open set of questions to comprehend figure captions, tables, and accompanying text \cite{Borchmann2021DUE}. Traditionally, the NLP community has focused on using \emph{clean} texts as context to their QA systems. However, this is not representative of the vast majority of scholarly information found in structured documents. As QA garners interest from the computer vision community, DocVQA \cite{mathew2021docvqa} and VisualMRC \cite{tanaka2021visualmrc} have extended document QA to extracting evidence from single images, paving the way to extend contexts from text to visual sources.

A foundational challenge in building robust document QA systems is ensuring well-formed contexts, which entails accurate text extraction and requires adaptation to new document layouts. Nonetheless, even when text can be cleanly extracted, there still remains the crucial task of identifying question-relevant paragraphs for answer prediction.

Our contribution is a \textbf{general-purpose system for QA on full documents in their original PDF form}, that addresses the key challenges of scientific document QA: (1) accurate text extraction from unseen layouts, (2) evidence retrieval (i.e., context selection), and (3) robust QA. A demo of our system is available through \href{https://huggingface.co/spaces/Epoching/DocumentQA}{Hugging Face}. 

\section{Dataset}
The Question Answering on Scientific Research Papers (\dataset) dataset consists of 1,585 NLP papers sourced from arXiv, and is accompanied by 5,049 questions from NLP readers and corresponding answers from NLP practitioners. Papers in \dataset are cited by their arXiv DOIs, which we used to fetch the original PDF documents as input to our system, as our work is focused on knowledge extraction at the PDF level. 

\dataset contains 7,993 answers categorized by answer type: \textit{Extractive} (4142), \textit{Abstractive} (1931), \textit{Yes/No} (1110), and \textit{Unanswerable} (810). Using only the \textit{Extractive}, \textit{Abstractive} and \textit{Yes/No} answers, we match our model prediction to the most similar answer when a question has more than one answer, and report our performance accordingly. 

\dataset is ideal for evaluating our proposed framework because it provides: (1) paragraph text and table information to evaluate our layout-analysis model (in its ability to cleanly extract document regions); (2) evidence paragraphs to validate, and optionally finetune, our evidence retrieval model (in its ability to retrieve good context paragraphs); and (3) ground-truth answers to assess the accuracy of our QA model (in its ability to answer the question given the context). 

\section{Methodology}
Document QA on raw PDFs is necessary towards automating knowledge extraction from scientific corpora and has remained an unaddressed problem. To address this, we propose a flexible information extraction tool to alleviate laboriously searching for answers grounded in evidence. Our system combines: (1) a robust text detector for visually rich documents, (2) explicit passage retrieval for evidence selection, and (3) multi-format answer prediction. We used pretrained open-source machine learning models that are effective in a zero-shot setting. We also finetuned these models to improve our system's end-to-end performance.

\subsection{Problem Description}

Our work addresses evidence retrieval at the PDF level. Thus, our document QA task is defined as: given a question and a PDF document, predict the answer to the question. We decompose this problem into three subtasks: text extraction (\sectionsymb~\ref*{subsec:extractor}), evidence retrieval (\sectionsymb~\ref*{subsec:retriever}), and QA (\sectionsymb~\ref*{subsec:qa}). 

First, the PDF document, represented as a series of images, has its semantic regions identified and their corresponding text content extracted as passages. Second, the passages are ranked by their relevance to the question. Irrelevant passages are filtered out so only the most relevant passages are used as contexts for QA. Finally, given a context and question, the answer is predicted. The overall architecture is shown in Figure~\ref{fig:system_arch}. These components correspond to the respective tasks of \textit{Detect}, \textit{Retrieve} and \textit{Comprehend}, or \drc, which is also the name of our proposed system.

\subsection{Detect}
\label{subsec:extractor}
The first step of our pipeline is to extract text from PDF documents. Libraries such as \pdfminer \cite{Shinyama2020PDFminer} and \tesseract \cite{Smith2007OCR} extract text from documents indiscriminately, including unwanted page numbers and footnotes, which would need to be filtered out before the extracted text can be used as context paragraphs. Thus, prior to text extraction, document layout analysis should be performed to detect targeted regions (from which text is to be extracted).

Document layout analysis models are trained to segment a document into its constituent components (e.g., paragraphs, figures, and tables). The Document Image Transformer (\dit) \cite{li2022dit} is designed for layout analysis and text detection. \dit uses a masked image modeling objective to pretrain a Vision Transformer \cite{dosovitskiy2021ViT} without labels. It supports prediction of semantic region bounding boxes and segmentation masks. Predicted regions are then passed to OCR tools for text extraction. 

In the \textit{Detect} stage of \drc, the \pdftoimage library first converts each page of the document to images. For each image, \dit detects the bounding boxes for paragraphs. The text within each bounding box is then extracted using \pdfminer. The extracted texts form passages which are candidates for question evidence. 

\subsection{Retrieve}
\label{subsec:retriever}
Evidence retrieval identifies relevant passages by ranking them according to their similarity to the question. We considered several architectures. 
\paragraph{Lexical Retriever}
\bestmatch \cite{Robertson2009BM25} ranks questions and passages based on token-matching between sparse representations of the question and passage. Prior work has shown that \bestmatch is a strong baseline across many datasets \cite{thakur2021beir, sciavolino2021}. Given a question $q$ containing tokens $q_1,\ldots,q_T$ and a set of passages $\set{P}$, the \bestmatch retrieval score $s$ between $q$ and passage $p \in \set{P}$ is defined using \textsc{TF-IDF} token weights:
\begin{equation*}
    s_{q,p}^{BM25} = 
    \sum_{i=1}^{T} \log\Bigl( \frac{|\set{P}|}{N(q_i, \set{P})}\Bigr)\frac{n(q_i,p)(k_1+1)}{k_1\bigl(1-b+\frac{b|p|}{avpl}\bigr) + n(q_i,p)}
\end{equation*}
where $|\set{P}|$ is the number of passages in the corpus; $|p|$ is the length of the passage; $N(q_i,\set{P})$ is the number of passages with token $q_i$; $n(q_i,p)$ is the term frequency of $q_i$ in passage $p$; $avpl$ is the average passage length. $k_1$ and $b$ are constants. 

\paragraph{Dual Encoder}
The Dense Passage Retriever (\dpr) \cite{karpukhin2020DPR} learns via a contrastive training objective with in-batch negatives and hard negatives chosen by \bestmatch. For question $q$ and a set of passages $\set{P}$, \dpr measures question-passage similarity with a dual-encoder architecture, where $f_q$ encodes the question $q$ and $f_p$ encodes the passage $p \in \set{P}$ to the same latent space \cite{Bromley1993Dual,Singh2021}. The retrieval score $s$ is defined as the dot product of the two resulting embeddings:
\begin{equation*}
    s_{q,p; \Phi}^{DE} = f_q(q; \Phi_q)^{\intercal} f_p(p; \Phi_p)
\end{equation*}
$\Phi=[\Phi_q,\Phi_p]$ denotes the retriever question and passage encoder parameters. We used the DPR \emph{multi} variant, which has been trained on additional data, as \citet{karpukhin2020DPR} has shown that the additional data improves retrieval generalizability. 

\paragraph{Cross-Encoder}
Instead of embedding the question $q$ and passage $p$ separately, cross-encoders \cite{Nogueira2019PassageRW} compute a retrieval score $s$ where $f(q,p;\Phi)_{[\text{CLS}]}$ encodes both question and passage using the CLS token representation of their concatenation:
\begin{equation*}
    s_{q,p;\Phi}^{CE} = \text{softmax}\Bigl(f(q,p;\Phi)_{[\text{CLS}]}W + b\Bigr)
\end{equation*}
where $W$ and $b$ are the weight and bias in the final layer that classifies whether $p$ is relevant to $q$. Many cross-encoders $f$ have since been proposed and a comparative analysis was performed \cite{Zhang2021Comp}, where the \electra{-base} \cite{clark2020electra} cross-encoder (\electrace) was declared as the best cross-encoder due to its stability and effectiveness across datasets. Thus, the \electrace model (trained on MS MARCO~\cite{bajaj2016ms}) was selected as our starting cross-encoder.

Once the passages have been ranked, the top-$K$ most relevant passages are used as contexts in the QA stage.

\begin{table}[t]
\begin{center}
\begin{tabular}{ll|ccc}
\hline\Tstrut{}\Bstrut{}
Type & Model & Epochs & WD & BS\\
\hline\Tstrut{}\Bstrut{}
& \bestmatch & -- & -- & -- \\
Retriever & \dprft & 20 &  0 & 8 \\
& \electraceft & 6 & 0.01 & 8 \\
\hline\Tstrut{}\Bstrut{}
Reader & \unifiedqaft & 10 & 0.01 & 10 \\
\hline
\end{tabular}
\caption{Tuned hyperparameter values for the number of epochs, weight decay (WD), and batch size (BS) for finetuning. The learning rate for all trainable models is 2e-5.
}
\label{tab:retriever_reader_params}
\end{center}
\end{table} 

\subsection{Comprehend}
\label{subsec:qa}
The final stage of \drc is comprehending a document's contents via  \textit{multi-format} question answering. These formats correspond to answer types, which can be extractive, abstractive, or Boolean. (An extractive answer is a span of text taken verbatim from an evidence passage. An abstractive answer is a generated span not quoted verbatim from the evidence. A Boolean answer is a binary prediction: yes or no.)

For comprehension, we use \unifiedqa \cite{khashabi2022unifiedqa}, a generative question-answering model that has been pretrained on 20 datasets and can predict all answer formats with a single architecture. The answer type returned by \unifiedqa depends on the way the question is phrased. For each of the $K$ relevant passages (from the \textit{Retrieve} stage), we pair the passage with the question as input to \unifiedqa to predict an answer. At the end, we have $K$ answers -- one for each of the $K$ passages.  

\begin{table*}[t]
    \begin{center}
    \begin{tabular}{l l l | c c c c c c c c}
        \toprule
        & & & \multicolumn{2}{c}{Extractive} & \multicolumn{2}{c}{Abstractive} & \multicolumn{2}{c}{Boolean} & \multicolumn{2}{c}{Overall} \\
        Extractor & Retriever & QA & Val. & Test & Val. & Test & Val. & Test & Val. & Test \\
        \midrule
        \tesseract & \electrace & \unifiedqa & 34.47 & 34.77 & 21.05 & 21.99 & 74.69 & 73.06 & 33.82 & 34.75 \\
        \ditt\,+\,\pdfminer& \electrace & \unifiedqa & 33.91 & 34.65 & 21.80 & 22.18 & 74.97 & 78.73 & 34.05 & 35.28 \\
        \ditt\,+\,\pdfminer & \electrace & \unifiedqaft & 38.46 & 39.70 & 25.23 & 24.24 & \textbf{84.95} & \textbf{85.29} & 38.81 & 39.22 \\
        \hdashline[.4pt/1pt] 
        \ditt\,+\,\pdfminer & \electraceft & \unifiedqa & 33.70 & 35.11 & 22.89 & 22.74 & 77.28 & 79.90 & 34.63 & 35.88 \\
        \ditt\,+\,\pdfminer & \electraceft & \unifiedqaft & \textbf{39.18} & \textbf{41.79} & \textbf{26.29} & \textbf{25.49} & 84.02 & 81.82 & \textbf{39.46} & \textbf{40.16}  \\
        \hdashline[.4pt/1pt]
        -- & -- &  \led{-base}            & 28.10 & 32.50     & 16.70 & 14.91     & 61.82 & 69.05     & 28.94 & 32.97 \\
        -- & -- &  \led{-base-scaff} & 23.37 & 29.59     & 15.49 & 14.95     & 66.36 & 67.14     & 26.37 & 31.59 \\
        \bottomrule
    \end{tabular}
    \caption{\unifiedqa \answerf scores using the top ranked context from extracted PDF regions.}
    \label{tab:end-to-end}
    \end{center}
\end{table*} 

\begin{table}[t]
\begin{center}
\begin{tabular}{ll|ccc}
\hline\Tstrut{}\Bstrut{}
Category & Method & $P$ & $R$ & $F_{1}$ \\
\hline\Tstrut{}\Bstrut{}
& \ditt\,+\,\pdfminer & \textbf{68.28} & 86.91 & \textbf{75.34} \\
Paragraphs & \pdfminert & 48.95 & \textbf{90.31} & 62.54 \\
& \tesseract & 49.27 & 89.75 & 62.73 \\
\hline\Tstrut{}\Bstrut{}
& \ditt\,+\,\pdfminer & \textbf{67.72} & 82.87 & \textbf{70.81} \\
Tables & \pdfminert & 6.88 & 92.71 & 12.38 \\
& \tesseract & 7.31 & \textbf{96.40} & 13.13 \\
\hline
\end{tabular}
\caption{Token extraction from paragraphs and tables within all documents from \dataset{}. $P$ denotes precision and $R$ denotes recall.}
\label{tab:dit_perf}
\end{center}
\end{table} 

\section{Experimental Setup}
\subsection{\dataset Baselines}
\label{subsec:baseline}
Following \citet{Dasigi2021QASPER}, in our \dataset experiments, we use the Longformer-Encoder-Decoder (\led) \cite{beltagy2020longformer} as the baseline model for evidence retrieval and QA. This model uses a modification of self-attention from the Transformer architecture \cite{vaswani2017Transformer} to encode longer sequences more efficiently. To jointly answer questions and decide whether a context is relevant in providing answer evidence, \led optimizes a multi-task objective. In addition to answer generation, \led adds a classification head (termed \emph{evidence scaffold}) that operates over each paragraph to predict binary labels (evidence or non-evidence). Since we discarded unanswerable questions from \dataset, we retrain \led on the remaining questions and evaluate with and without evidence scaffolding. The retrained \led serves as a fairer competitor to \unifiedqa, which was not pretrained on unanswerable questions.

\subsection{Text Extraction with Layout Analysis}
We use \dit with \pdfminer for selective text extraction. First, a pretrained \dit model predicts the bounding boxes of paragraphs on each page. Then, \pdfminer extracts text within the bounding boxes. We denote this two-step procedure as \ditt+\pdfminer, and compare against \tesseract, which takes an image as input and returns the text found within the image, as well as \pdfminer's high-level extractor (\pdfminert), which takes a PDF as input and exploits PDF metadata to extract texts within the pages.
 
\subsection{Retriever-QA Implementation Details}
For \bestmatch, we create an inverted index on \dataset validation and test sets using Pyserini \cite{Lin2021Pyserini} with default parameters ($k_1$=0.9, $b$=0.4). For \dpr and \electrace, we start with pretrained models from Hugging Face, then finetune them per hyperparameters shown in Table~\ref{tab:retriever_reader_params}. In finetuning \dpr and \electrace, we sample batches containing a 1:4 ratio of positive to negative evidence passages.

For \unifiedqa, we use the unifiedqa-v2-t5-large-1363200 model from Hugging Face. We finetune it in a  weakly supervised manner using evidence passages ranked by \electrace but with the original questions and answers from \dataset. The choice to use retrieved passages (instead of the human-labeled evidence passages from \dataset) should make our system more robust to noisy context paragraphs. We show that a pretrained text extractor and evidence retriever can adapt \unifiedqa to the domain of \dataset papers without labeled evidence.

\begin{table*}[t!]
    \setlength\tabcolsep{5pt}
    \centering
    \begin{tabular}{ll|cccccccc}
    \toprule
     &  & \multicolumn{2}{c}{\textbf{\textit{K=1\%}}} & \multicolumn{2}{c}{\textbf{\textit{K=5\%}}} & \multicolumn{2}{c}{\textbf{\textit{K=10\%}}} & \multicolumn{2}{c}{\textbf{\textit{K=20\%}}} \\
    Retriever & Training & Val. & Test & Val. & Test & Val. & Test & Val. & Test \\
    \midrule
\bestmatch & -- & 15.32 & 15.97 & 33.00 & 31.81 & 45.36 & 44.28 & 61.95 & 60.38 \\
    \dpr & -- & 10.91 & 11.75 & 23.23 & 25.28 & 33.49 & 36.50 & 51.13 & 52.61 \\
    \electrace & -- & \underline{20.98} & \underline{22.82} & \underline{39.44}  & \underline{39.97} & \underline{52.08} & \underline{52.06} & \underline{66.89} & \underline{66.39} \\
    \midrule
    \dprft & \dataset & \textbf{25.21} & \textbf{26.22} & 43.79 & 46.41 & 58.19 & 59.09 & 73.31 & 72.97 \\
    \electraceft & \dataset & 22.81 & 24.50 & \textbf{49.26} & \textbf{50.36} & \textbf{64.58} & \textbf{65.67} & \textbf{78.80} & \textbf{79.66} \\
    \bottomrule
    \end{tabular}
    \caption{Retrieval recall measured as the top percentages (top-$K\%$) of retrieved passages that contain the answer, averaged across all questions. 
        The \textit{Training} column makes explicit which retrievers are applied zero-shot or finetuned using the \dataset train split. For each pairing of data split and $K$, the best performing model is shown in \textbf{bold} and the second best is \underline{underlined}.
}
    \label{tab:retr_perf}
\end{table*} 

\subsection{Evaluation Metrics}
We evaluate \drc's text extraction, retrieval, and QA stages separately. For each stage, performance is measured by the $F_1$ score between the predicted outputs and the target labeled in \dataset. Adopting the same notation as \citet{Dasigi2021QASPER}, we name the $F_1$ scores for our evidence retrieval and QA as \evidencef and \answerf, respectively. For text extraction, in addition to its $F_1$ score, we also evaluate its precision and recall.

Since each question in \dataset is labeled with its answer(s) and accompanying evidence, it is possible to evaluate both our QA and evidence retrieval stages using this single dataset. For QA, \answerf is calculated between the tokens in the predicted answer and the tokens in the target answer. For our evidence retrieval stage, which ranks passages by their relevance to a given question, \evidencef is calculated between a fixed percentage of the top ranking passages and the set of passages labeled as evidence in \dataset.

\dataset also contains the plain text for each of its documents, organized so that text from paragraphs and tables are separated. We use this plain text to evaluate the efficacy of our text extraction to extract only the primary content of PDF document. The precision, recall, and $F_1$ score for text extraction are calculated between the tokens in a document's extracted text and its tokens in \dataset's plain text version. For all of our experiments, tokenization is performed at the word level, using our pretrained \unifiedqa model's tokenizer.

\section{Results}
\label{sec:results}
We demonstrate \drc's effectiveness on document QA by measuring its end-to-end performance. We also evaluate its constituent components on text detection, evidence retrieval, and QA tasks against existing \dataset baselines. For evidence retrieval, we study the benefits of having a separate retrieval process, in contrast to the evidence selection scaffold for \led. Furthermore, we explore \drc's performance in both zero-shot and finetuned settings, to assess its performance under varying degrees of access to labeled data.

\begin{table*}[t]
    \begin{center}
    \begin{tabular}{l l | c c c c c c c c}
        \toprule
                        & & \multicolumn{2}{c}{Extractive} & \multicolumn{2}{c}{Abstractive} & \multicolumn{2}{c}{Boolean} & \multicolumn{2}{c}{Overall} \\
        Retriever & QA & Val. & Test & Val. & Test & Val. & Test & Val. & Test \\
        \midrule
        \bestmatch & \unifiedqa & 29.24 & 30.04 & 23.12 & \underline{24.77} & \underline{77.40} & 78.99 & 33.23 & 36.85 \\
        \dpr & \unifiedqa & 28.28 & 28.94 & 23.60 & 21.85 & 75.96 & 79.38 & 32.58 & 35.59 \\
        \electrace & \unifiedqa & \underline{32.72} & \underline{34.46} & \underline{24.34} & 23.40 & 77.30 & \underline{80.63} & \underline{35.73} & \underline{39.51} \\
        \hdashline[.4pt/1pt]
        \dprft & \unifiedqa
            & \textbf{33.37} & \textbf{34.64} & \textbf{24.97} & 24.53 & \textbf{78.66} & 77.16 & \textbf{36.46} & 39.31 \\
        \electraceft & \unifiedqa
            & 33.18 & 34.21 & 24.56 & \textbf{25.78} & 78.20 & \textbf{81.45} & 36.18 & \textbf{40.03} \\
        \hdashline[.4pt/1pt]
        -- & \led{-base}            & 28.10 & 32.50     & 16.70 & 14.91     & 61.82 & 69.05     & 28.94 & 32.97 \\
        -- & \led{-base-scaff}  & 23.37 & 29.59     & 15.49 & 14.95     & 66.36 & 67.14     & 26.37 & 31.59 \\
\bottomrule
    \end{tabular}
    \caption{\unifiedqa Answer-$F_{1}$ scores on \unifiedqa{} using the top ranked context from selected retrievers compared to the \led baselines. For all data splits, the best performing model is shown in \textbf{bold} and the best zero-shot model is \underline{underlined}.}
    \label{tab:reader_perf_k1}
    \end{center}
\end{table*} 

\begin{table}[t]
\centering
\begin{tabular}{lcc}
\toprule
\multirow{2}{*}{Model} & \multicolumn{2}{c}{\evidencef{}}\\
& Val. & Test \\
\midrule
\electrace & \textbf{31.75} & 36.37\\
\electraceft & 31.58 & 36.12\\
\hdashline[.4pt/1pt]
\led{-base} & 23.94 & 29.85\\
\led{-base-InfoNCE} & 24.90 & 30.60 \\
\led{-large} & 31.25 & \textbf{39.37} \\
\bottomrule
\end{tabular}
\caption{Comparison between \electra cross-encoders against \led  baselines in terms of \evidencef.}
\label{tab:evidence_selection}
\end{table} 

\subsection{End-to-End QA System}
\label{subsec:results_e2e}
Table~\ref{tab:end-to-end} shows \drc's performance in terms of \answerf. In these experiments, we extract text from documents using either \tesseract or \dit and rank passages using \electrace. We then pass highly-ranked passages to \unifiedqa for answer prediction. First, we study the influence of the text detection model on \answerf performance by comparing \tesseract to \dit. While we observe that using \dit reports higher \answerf than  \tesseract across all answer types, the difference is negligible. 

Next, we examine \drc in the zero-shot setting where \electrace and \unifiedqa models are not finetuned. \drc achieves an overall +2.31 improvement in \answerf over \led{-base} without scaffolding on \dataset{'s} test split. To improve upon the fully zero-shot approach, we apply weak supervision to finetune \unifiedqa: we sample extracted passages according to their retrieval scores from the pretrained \electrace model, assuming that higher ranked passages are correlated with selection probability for answer prediction. Thus, we are able to finetune \unifiedqa without access to human-labeled contexts, since labeled question-answer pairs are generally unavailable for large technical corpora. This finetuning approach yields a +6.25 improvement to \led{-base} in overall \answerf on the test dataset. 

To analyze \answerf performance when ground-truth question-passage pairs are available, we consider an \electrace retriever finetuned on \dataset{'s} training set. We then finetune \unifiedqa through weak supervision using the now improved retriever. \drc with a finetuned \electrace shows modest gains over the zero-shot system but still lesser performance compared to the pretrained \electrace with a weakly-supervised \unifiedqa. This suggests that downstream QA performance is better improved by adapting to the target domain \dataset documents, than by receiving more relevant passages.

Combining a finetuned \electrace retriever with a weakly supervised \unifiedqa model shows the greatest improvements over \led{-base} without scaffolding, +7.19 in \answerf on the and test dataset for all answer types. We observe that using finetuned \electrace for weak supervision shows worse performance on Boolean questions than using the pretrained \electrace to weakly supervise \unifiedqa. This discrepancy is likely due to the small proportion of Boolean samples in the validation and test datasets compared to other formats, 13\% and 15\% respectively. 

Across all experiments, \drc demonstrates superior performance to \led while solving a more difficult task: \drc starts from PDFs while \led starts from clean texts. \drc bridges an essential gap in real-world applications for scientific knowledge extraction because PDFs are directly processed as input. In the following discussion, we validate our individual system components.

\subsection{Text Detection}
\label{subsec:text_detection}
We evaluate three different methods for text extraction from PDF files: (1) \ditt+\pdfminer, (2) \pdfminert, and (3) \tesseract. Table~\ref{tab:dit_perf} reports the average precision, recall, and $F_1$ between the extracted tokens and those in the ground-truth text.

For paragraph extraction, \ditt+\pdfminer has better precision than \pdfminert~(+19.33) and \tesseract (+19.01). We attribute this improvement to extracting fewer unwanted artifacts (e.g., page numbers, headers, footers, and footnotes). For text within tables, only \ditt+\pdfminer is effective off-the-shelf. \pdfminert~and \tesseract do not disambiguate between text in and outside of tables. 
\pdfminert~and \tesseract would suffice if text is contained only in tables, or only in paragraphs, but not a mixture of the two because the text from tables and paragraphs will be interspersed.

\subsection{Evidence Passage Retrieval}
\label{subsec:results_retrieval}
We compare \dpr{}, \electrace{}, and \bestmatch{} by their ability to rank passages by relevance to questions. Table~\ref{tab:retr_perf} shows the recall of evidence passages within various percentages of the top ranked passages, averaged over all questions in \dataset, for the retrievers in both zero-shot and finetuned settings. As questions are posed for a specific document, our retrievers consider a variable number of passages per question because documents vary in length. Since top-$K$ penalizes longer documents when $K$ is small, we measure recall using top-$K$\% for $K \in \{1, 5, 10, 20\}$. 

In the zero-shot setting, \bestmatch outperforms \dpr by an average of +15.58 gain in recall on the test data. These results support findings from \citet{sciavolino2021}, who reported that \dpr trained on Natural Questions \cite{Kwiatkowski2019NQ} underperformed \bestmatch when faced with the new question patterns and entities found in their EntityQuestions dataset. Thus, \dpr requires finetuning and is less generalizable than \bestmatch, which has no trainable parameters. 

\electrace, with average recall gains of +7.2 and +13.78 over \bestmatch and \dpr, respectively, is the clear winner. We hypothesize that \electrace's success is due to the explicit interaction between every token of the question and passage through its cross-attention mechanism, offering a more expressive similarity function than \dpr{'s} inner product between question and passage or \bestmatch{'s} weighted term matching. 

To analyze how retrievers perform with ground-truth question-passage pairs, we also evaluate passage retrieval with \dpr and \electra cross-encoders finetuned on \dataset. Here, \electrace outperforms \dpr on the test data for $K$ = 5\%, 10\% and 20\% by an average recall of +3.95, +6.58, and +6.69, respectively. Notably, \dpr has higher recall for $K$=1\%. We conjecture that this may be due to \dpr's contrastive objective utilizing hard negative sampling, but further analysis on the relationship between training objective and ranking is needed.

\subsubsection{Comparison to Evidence Selection Scaffold}
\label{subsec:results_scaffold}

To compare against \led{'s} evidence scaffold, we now treat \electrace as a binary classifier. Akin to \led{'s} evidence scaffold, we use the [CLS] representation of the question-passage pair as input to a single layer neural network to estimate the probability that the passage is relevant as evidence to the question and use a classification threshold of 0.5 \cite{Nogueira2019PassageRW}. Table~\ref{tab:evidence_selection} illustrates the evidence classification performance of zero-shot and finetuned \electrace models against \led variations. \evidencef scores are computed using the extracted passages classified as evidence with respect to the ground-truth set labeled in \dataset. We observe that the difference between the zero-shot and finetuned \electrace models is negligible. On the test split, zero-shot \electrace shows a notable \evidencef improvement over \led-base augmented with InfoNCE loss \cite{caciularu2021}, but is outperformed by \led{-large}. This agrees with findings from \citet{Dasigi2021QASPER} that \led{-large} generally outperforms \led{-base} for retrieval but not QA. Thus, we consider only \led{-base} in our subsequent experiments with downstream QA.

\subsection{Question Answering}
\label{subsec:results_qa}

Here, we focus on the effect of using retrieval mechanisms on \dataset{'s} plain text passages for answer prediction. We report \answerf scores for extractive, abstractive, and Boolean answer types. We also finetune \dpr and \electrace models on \dataset{'s} train split 
and compare against \led variations. 

Table~\ref{tab:reader_perf_k1} shows \unifiedqa{'s} \answerf using the highest-ranked passage from each retriever. We observe that \unifiedqa (first 5 rows) generally yields higher \answerf scores across answer types, datasets, and retrievers than \led baselines (last 2 rows). The exception is extractive answers from the test set where \bestmatch reports a lower \answerf than \led{-base}. Similarly, a zero-shot \dpr retriever performs worse than both \led models. Among zero-shot retrievers, \electrace yields the best performance on the overall test set with a +7.92 and +6.54 \answerf increase over \led with and without scaffolding. While \dpr benefits the most from finetuning, the finetuned \electrace reports the highest test performance across all answer types.

\begin{figure}[t]
\centering
\hspace*{-9mm}\includegraphics[width=0.9\columnwidth]{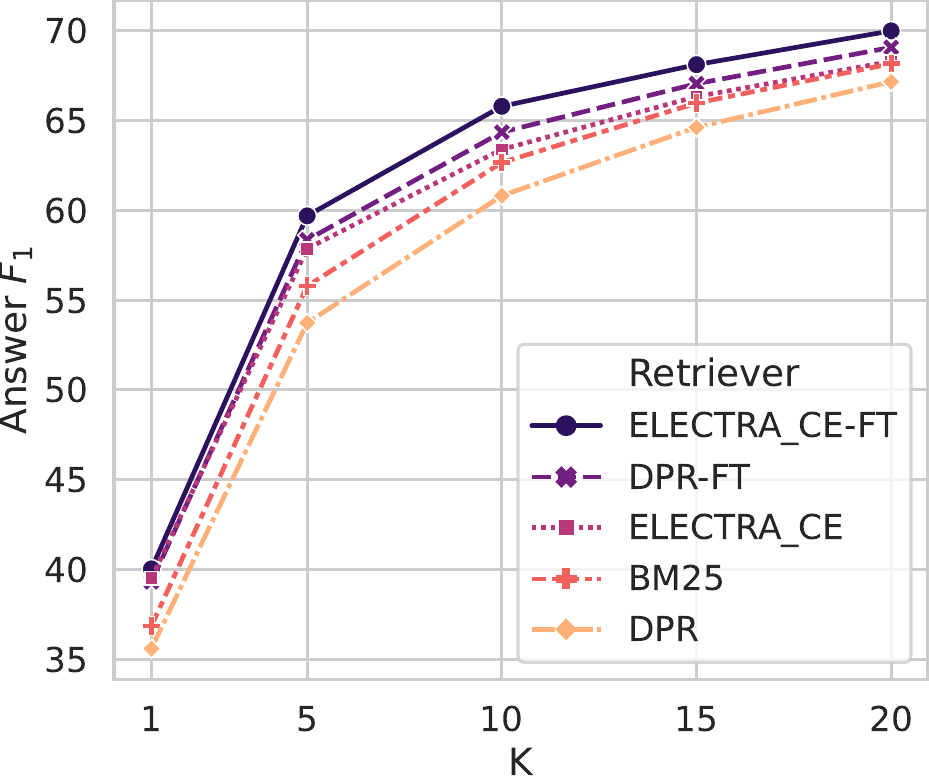} 
\caption{UnifiedQA overall \answerf performance on the \dataset{} test data split using the top-$K$ ranked contexts from selected retrievers. Retrievers with ``-FT'' denote those 
finetuned on \dataset{}'s train split.}
\label{fig:reader_k>1}
\end{figure}

To verify that a weakly-supervised \unifiedqa model improves end-to-end answer prediction using weakly-labeled evidence, we measure the effect of retrieval beyond the top-ranked passage. We report \unifiedqa{'s} overall \answerf on the test data using the best answer predicted from the top-$K$ ranked passages. Figure~\ref{fig:reader_k>1} compares performance using zero-shot and finetuned retrievers. Zero-shot \electrace consistently dominates \bestmatch and \dpr without finetuning. When finetuned, \electrace reports the best performance for all choices of $K$. This confirms our hypothesis that we can adapt \unifiedqa to achieve higher quality answers, via weak supervision using ranking signals from the retriever.

\section{Conclusion}
We introduced \drc, an end-to-end QA system for automating manual knowledge extraction from scientific PDF documents. We showed that \drc greatly improves over existing baselines, which act on clean texts and sidestep the challenge of PDF-to-text extraction. Through extensive experiments, we evaluate our pipeline components in both zero-shot and finetuned settings. In practice, datasets as comprehensive as \dataset are few and may not be feasible for niche domains. In such cases, a fully zero-shot pipeline is mandatory for document QA, and \drc can be weakly supervised to adapt to specific domains. Our \drc sets a new benchmark for \dataset and serves as a proof of concept for an end-to-end document QA system, from PDF to answer. Key takeaways from our experiments include:

\begin{enumerate}
    \item \dit demonstrates superior text extraction performance to \pdfminer and \tesseract.
    
    \item Zero-shot \electrace offers the best retrieval performance for all top-$K\%$ (where $K \in \{1, 5, 10, 20\}$).

    \item \drc adapts to new domains through weakly-supervised training on evidence passages leading to substantially improved answer prediction over \led baselines.
\end{enumerate}
A demo of \drc is available through \href{https://huggingface.co/spaces/Epoching/DocumentQA}{Hugging Face}. 

In this work, we have only scratched the surface with text. Future QA systems should be able to process scientific documents with diverse layouts and visually rich content. In order to fully automate information extraction beyond text, we must augment our system to identify and understand visual elements (e.g., figures) by incorporating visual question answering (VQA) and multimodal representations. Additionally, \dataset consists of well-formatted research papers in digitally-generated PDF documents. Further experiments are required to evaluate \drc{'s} performance under domain shifts with respect to document format and text cleanliness (e.g., OCR noise).

Deployed document QA systems must operate on documents of varying lengths. We demonstrated retrieval and evidence selection on relatively short research papers, whose lengths are unrepresentative of manual processing tasks that require comprehending entire textbooks and technical manuals. As the number of passages grows, the input sequence limit will be exceeded, making answer prediction and evidence selection by efficient transformer architectures challenging. In addition, ranking via cross-encoder may be computationally expensive due to the costly cross-attention operation between the question and each passage. Thus, QA on lengthy documents may require a dual-encoder retriever, to store precomputed passage embeddings using FAISS \cite{johnson2019billion} to maximize efficiency. 

\section*{Acknowledgements}
This work was performed under the auspices of the U.S. Department of Energy by Lawrence Livermore National Laboratory under Contract DE-AC52-07NA27344. 

\bibliography{aaai23}

\end{document}